\documentclass[10pt,twocolumn,letterpaper]{article}

\usepackage[table,xcdraw]{xcolor}
\usepackage{comment}
\usepackage{booktabs}
\usepackage{multirow}
\usepackage{mathtools}
\usepackage{float}
\usepackage{graphicx}
\usepackage[accsupp]{axessibility}

\usepackage{iccv}              

\newcommand{\mymodel}{\textsc{GDKVM}\xspace}

\definecolor{iccvblue}{rgb}{0.21,0.49,0.74}
\usepackage[pagebackref,breaklinks,colorlinks,allcolors=iccvblue]{hyperref}

\def\paperID{9792} 
\def\confName{ICCV}
\def\confYear{2025}

\title{\mymodel: Echocardiography Video Segmentation via \\ Spatiotemporal Key-Value Memory with Gated Delta Rule}

\author{
    Rui Wang\textsuperscript{\rm 1}~
    Yimu Sun\textsuperscript{\rm 1}~
    Jingxing Guo\textsuperscript{\rm 1}~
    Huisi Wu\textsuperscript{\rm 1}\thanks{Corresponding Author}~
    Jing Qin\textsuperscript{\rm 2}
\\
    \small{\textsuperscript{\rm 1}College of Computer Science and Software Engineering, Shenzhen University}\\
    \small{\textsuperscript{\rm 2}Centre for Smart Health, School of Nursing, The Hong Kong Polytechnic University}\\
    {\tt\scriptsize 2400101058@mails.szu.edu.cn, hswu@szu.edu.cn}
}

\begin{document}

\maketitle
\begin{abstract}
Accurate segmentation of cardiac chambers in echocardiography sequences is crucial for the quantitative analysis of cardiac function, aiding in clinical diagnosis and treatment.
The imaging noise, artifacts, and the deformation and motion of the heart pose challenges to segmentation algorithms. 
While existing methods based on convolutional neural networks, Transformers and space-time memory networks, have improved segmentation accuracy, they often struggle with the trade-off between capturing long-range spatiotemporal dependencies and maintaining computational efficiency with fine-grained feature representation. 
In this paper, we introduce \textbf{GDKVM}, a novel architecture for echocardiography video segmentation. 
The model employs Linear Key-Value Association (LKVA) to effectively model inter-frame correlations, and introduces Gated Delta Rule (GDR) to efficiently store intermediate memory states. Key-Pixel Feature Fusion (KPFF) module is designed to integrate local and global features at multiple scales, enhancing robustness against boundary blurring and noise interference. 
We validated GDKVM on two mainstream echocardiography video datasets (CAMUS and EchoNet-Dynamic) and compared it with various state-of-the-art methods. 
Experimental results show that GDKVM outperforms existing approaches in terms of segmentation accuracy and robustness, while ensuring real-time performance. 
Codes are available at \url{https://github.com/wangrui2025/GDKVM}. 
\end{abstract}    
\section{Introduction}
    \label{sec:intro}
Echocardiography is a vital imaging modality for cardiac assessment, valued for its non-invasive nature and real-time imaging capabilities. Accurate segmentation of cardiac chambers from echocardiography videos is a foundational step for quantitative analysis~\cite{el2022semi}. Delineating the left ventricle is crucial for calculating the left ventricular ejection fraction, a cornerstone metric for diagnosing cardiovascular diseases and guiding treatment strategies \cite{amer2021resdunet,yang2024echocardiography}.
                \begin{figure}[t]
    \centering
    \begin{subfigure}{0.163\linewidth}
        \includegraphics[width=\linewidth]{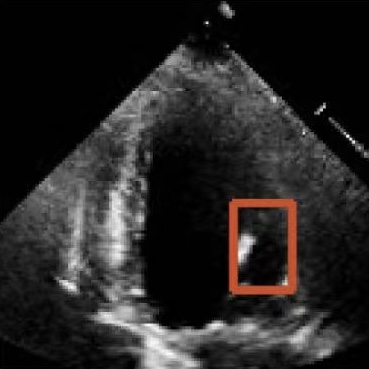}
        \caption{}
        \label{fig:cha:1}
    \end{subfigure}\hfill %
    \begin{subfigure}{0.163\linewidth}
        \includegraphics[width=\linewidth]{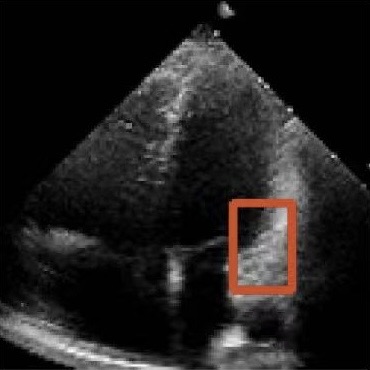}
        \caption{}
        \label{fig:cha:2}
    \end{subfigure}\hfill %
    \begin{subfigure}{0.163\linewidth}
        \includegraphics[width=\linewidth]{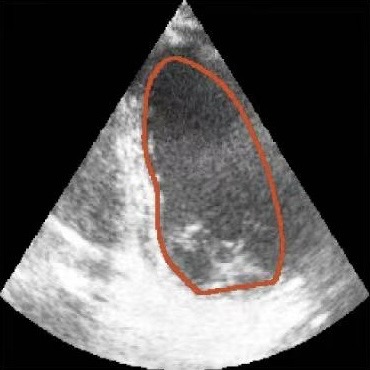}
        \caption{}
        \label{fig:cha:3}
    \end{subfigure}\hfill %
    \begin{subfigure}{0.163\linewidth}
        \includegraphics[width=\linewidth]{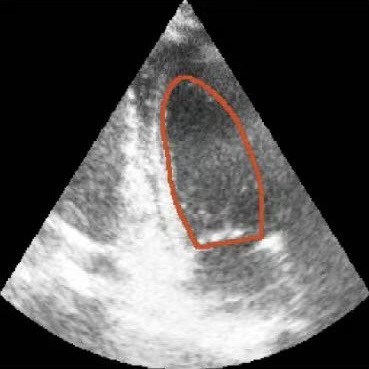}
        \caption{}
        \label{fig:cha:4}
    \end{subfigure}\hfill %
    \begin{subfigure}{0.163\linewidth}
        \includegraphics[width=\linewidth]{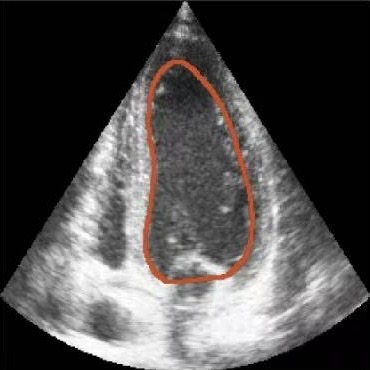}
        \caption{}
        \label{fig:cha:5}
    \end{subfigure}\hfill %
    \begin{subfigure}{0.163\linewidth}
        \includegraphics[width=\linewidth]{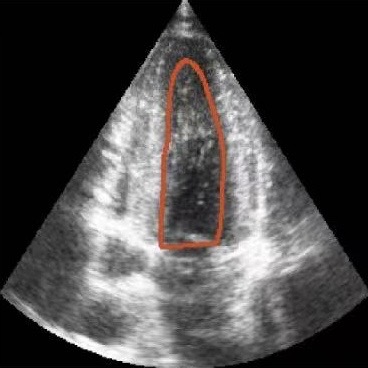}
        \caption{}
        \label{fig:cha:6}
    \end{subfigure}
    \caption{Illustrative challenges for echocardiography video segmentation: (a) speckle noise, (b) indistinct or blurred contours, and (c-f) the substantial changes in the target's shape and scale throughout the cardiac cycle.}
    \label{fig:cha}
\end{figure}

However, achieving precise segmentation in echocardiography faces severe challenges stemming from poor image quality and complex cardiac dynamics. \Cref{fig:cha:1,fig:cha:2} show that ultrasound images are characterized by high speckle noise and low contrast, which obscure tissue structures and lead to weak or incomplete anatomical boundaries. These artifacts hinder the model's ability to learn robust features and can result in the inaccurate segmentation. In the temporal dimension, the heart undergoes significant non-rigid deformation throughout the cardiac cycle. \Cref{fig:cha:3,fig:cha:4,fig:cha:5,fig:cha:6} show that, the shape and scale of the LV change dramatically between systole and diastole. This substantial dynamic variation demands that segmentation models possess strong temporal modeling capabilities to accurately track the changing appearance of the target across the video sequence~\cite{li2024automatic}.

Convolutional Neural Networks~\cite{oshea2015introductionconvolutionalneuralnetworks}, particularly U-Net architectures \cite{ronneberger2015u,long2015fully,iglovikov2018ternausnet}, marked a significant milestone in echocardiography segmentation. Their proficiency in learning hierarchical spatial features improved delineation robustness against noise~\cite{jha2019resunet++}. Yet, their inherently local receptive fields limit performance where boundaries are ambiguous and preclude the direct modeling of inter-frame temporal dependencies~\cite{painchaud2022echocardiography}. Vision Transformers address the local-context limitation by providing a global field of view \cite{wang2025transformer,liao2023left}, but their quadratic computational complexity and large data requirements present practical barriers. Spatiotemporal models incorporating recurrent units like ConvLSTM~\cite{chen2021assessing} aim for temporal coherence, yet they risk propagating errors through sequences and add significant computational overhead. A gap thus remains for a model that can integrate global temporal context efficiently while preserving local spatial precision.

This paper introduces \textbf{\mymodel}, a new architecture for echocardiography video segmentation designed to synthesize global temporal modeling with local feature precision in a computationally efficient manner. The core of \mymodel is a Linear Key-Value Association (LKVA) mechanism that captures inter-frame relationships. 
To efficiently update and preserve essential temporal information, we propose Gated Delta Rule (GDR) module that discards irrelevant historical memory while retaining features pertinent to the current frame. 
Key-Pixel Feature Fusion (KPFF) module enhances spatiotemporal representation by integrating local key features, global key features, and pixel-level features.
We evaluate \mymodel on two widely adopted echocardiography video datasets, CAMUS~\cite{leclerc2019deep} and EchoNet-Dynamic~\cite{ouyang2019echonet}, and compare it with various task-specific state-of-the-art methods as well as recent STM models. Experimental results indicate that \mymodel achieves consistently higher segmentation accuracy than existing approaches, demonstrating its effectiveness and robustness on both datasets.
Our main contributions are summarized as follows:
\begin{itemize}
  \item We propose a novel model \mymodel for echocardiography video segmentation, which fully leverages the powerful representational capacity of linear key-value association with new adaptations to ultrasound videos.
  \item We introduce Gated Delta Rule (GDR) and Key-Pixel Feature Fusion (KPFF) module; GDR employs interactions between the current frame and memory bias, helping the model manage its memory. KPFF enables the model to acquire global spatiotemporal features. 
  \item We conduct extensive experiments on the CAMUS and EchoNet-Dynamic datasets to validate the superior performance of our approach over current state-of-the-art methods.
\end{itemize}
\section{Related Work}
    \label{sec:related}
            \begin{figure*}[!t]
    \centering
    \includegraphics[width=0.95\textwidth]{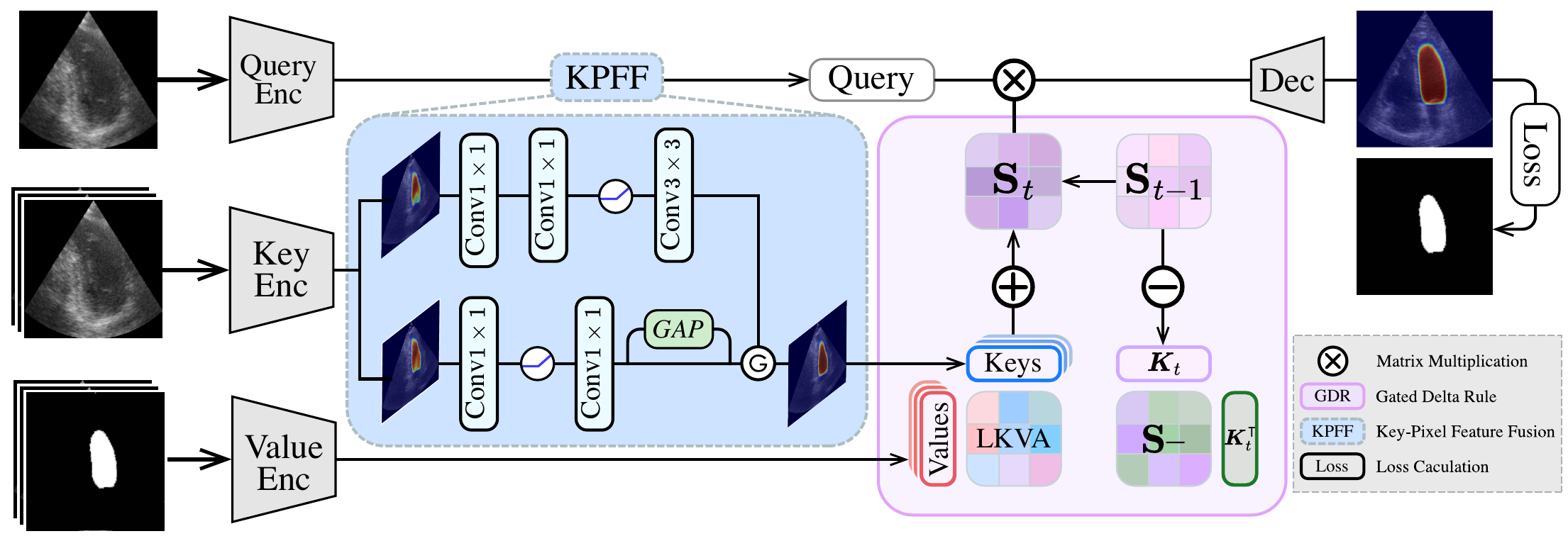}
    \caption{An illustration of \mymodel architecture. Linear Key-Value Association defines frame-to-frame causal relations as the state transition matrix. Gated Delta Rule helps in dynamically managing memory. Key-Pixel Feature Fusion fuses the local key feature, the global key feature with the pixel feature.}
    \label{fig:met:gdkvm}
\end{figure*}
    \paragraph{Echocardiography Video Segmentation.}
    \label{related:evs}
Echocardiography video segmentation aims for the temporally coherent segmentation of key cardiac structures, such as the left ventricle, which is crucial for the accurate assessment of cardiac functions like ejection fraction. Fully-supervised methods exhibit excellent performance, but they rely on dense, frame-by-frame annotation that is both time-consuming and costly. Sparsely-supervised methods, which require only a few annotated frames, have thus become an active area of research, aiming to effectively balance segmentation accuracy with annotation overhead. 
Multi-frame aggregation methods~\cite{sokeh2018superframes, liu2018path} and Space-Time Memory Networks (STM)~\cite{oh2019videoobjectsegmentationusing} are widely used in video segmentation tasks to leverage temporal information. The former processes multiple frames simultaneously to improve segmentation performance in long sequences, but demands substantial GPU resources in large-scale scenarios. By storing the features and segmentation results of past frames in a memory bank and retrieving only the necessary semantic information when a new frame arrives, STM networks significantly reduce memory overhead while maintaining temporal consistency. Methods such as XMem~\cite{cheng2022xmemlongtermvideoobject} and XMem++~\cite{bekuzarov2023xmemproductionlevelvideosegmentation} introduce multi-level memory mechanisms, drawing on human cognitive processes to sustain effective segmentation in extremely long videos while controlling memory usage.
Existing Space-Time Memory Networks \cite{oh2019videoobjectsegmentationusing, cheng2021rethinkingspacetimenetworksimproved, cheng2022xmemlongtermvideoobject, bekuzarov2023xmemproductionlevelvideosegmentation, cheng2024puttingobjectvideoobject} typically rely on mask propagation from fully annotated reference frames, yet they lack functionality to identify phases of the cardiac cycle. In medical applications, where it may be necessary to automatically locate end-systole and end-diastole frames, this limitation can reduce segmentation accuracy at critical moments. MemSAM~\cite{deng2024memsam} incorporates richer temporal cues into the memory bank, reducing misinterpretation of ambiguous structures and mitigating error propagation, but its computational overhead remains high, hindering direct deployment in clinical settings. Consequently, achieving both low computational cost and precise cardiac cycle recognition is a critical challenge that spatiotemporal memory methods urgently need to address in medical video segmentation.
    
\noindent
\textbf{Linear Attention Methods.}
    \label{related:linear}
Traditional self-attention~\cite{vaswani2017attention} exhibits quadratic growth in computational complexity as sequence length increases, leading to enormous resource overhead. 
To address this issue, Linear Attention approximates the original dot-product similarity calculation by introducing techniques such as a kernel function \cite{katharopoulos_transformers_2020, choromanski2020rethinking} or by employing low-rank decomposition~\cite{wang2020linformer}. By changing the order of computation or using matrix decomposition, these methods avoid the explicit construction of the attention matrix, which has a quadratic complexity.
This approach significantly reduces the computational and memory burden of Transformers for long-sequence tasks. 
Vision LSTM~\cite{beck:24xlstm}, a linear RNN methods, uses a stacked LSTM structure to extract multi-scale visual features and integrates temporal information through gating mechanisms, achieving strong performance in tasks such as video understanding. Mamba~\cite{gu2023mamba} and its variant Video Mamba~\cite{li2024videomambastatespacemodel}, based on state-space models (SSMs), efficiently model videos and capture rich spatiotemporal dependencies in long-sequence scenarios, demonstrating excellent scalability and expressive power.
However, these methods have not yet been thoroughly validated in the field of medical ultrasound imaging. This domain is particularly challenging due to its complex noise and low contrast. 
Optimizing model structures and training strategies for medical scenarios remains key to further unleashing the potential of linear attention.
\section{Method}
    \label{sec:method}
    \subsection{Overview}
    \label{subsec:overview}
We propose \textbf{\mymodel}, a linear key-value memory network designed to address the challenges of high computational cost and poor noise robustness in echocardiography video segmentation. 
The architecture is illustrated in \cref{fig:met:gdkvm}.

The key encoder uses the ResNet-50~\cite{he2015deepresiduallearningimage} backbone to extract raw features. 
The KPFF module integrates the fused features obtained by combining key features, global key features, and pixel features to get the $\mathbf{Keys}$. 
The value encoder encodes the original frame and the previous prediction mask into the $\mathbf{Values}$. We combine $\mathbf{Keys}$ and $\mathbf{Values}$ into a recurrent hidden $\mathbf{State}$.
During the linear update of the state, we use 2 data-dependent matrices, $\boldsymbol{\alpha}_t$ and $\boldsymbol{\beta}_t$, to control the state decay, both of which are projected from the previous state $\mathbf{S}_{t-1}$. $\boldsymbol{\beta}_t$ balances how much new information is incorporated into the state, while $\boldsymbol{\alpha}_t$ adaptively forgets old memories.
Using the $\mathbf{Query}$ frame, the readout is derived from the state $\mathbf{S}_t$, and the predicted segmentation result is obtained through the Decoder.
To simulate a clinical scenario, the model does not have access to the ground truth when making a prediction. During each training step, the model first predicts the morphology of the first and last frames; only then is the loss calculated by comparing the prediction against the ground truth~\cite{deng2024memsam}.
    \subsection{Linear Key-Value Association}
    \label{subsec:lkva}
\paragraph{Softmax matching.}
The input frames of an echocardiography video can be described as $\boldsymbol{I}_{1:t} \in \mathbb{R}^{tHW \times C_d}$. Current Transformers and Space-Time Memory Networks \cite{cheng2021rethinkingspacetimenetworksimproved,cheng2022xmemlongtermvideoobject,bekuzarov2023xmemproductionlevelvideosegmentation,cheng2024puttingobjectvideoobject} use softmax attention matching to maintain the memory bank for segmenting the object:
\begin{equation}
    \begin{aligned}
		\boldsymbol{O}_t &= \sum_{i=1}^{t} \frac{ \exp(\boldsymbol{K}_i^{\intercal} \boldsymbol{Q}_t)}{\sum_{j=1}^{t} \exp(\boldsymbol{K}_j^{\intercal} \boldsymbol{Q}_t)} \boldsymbol{V}_i,
    \end{aligned}
	\label{eq:softmax_attention}
\end{equation}
where $\boldsymbol{Q}_t, \boldsymbol{K}_t, \boldsymbol{V}_t \in \mathbb{R}^{tHW \times C_d}$. 
The $\mathcal{O}(t^{2} C_d)$ computational cost required by attention significantly increases inference time and creates significant pressure for real-time application requirements.
\paragraph{Linear matching.}
Linear attention \cite{katharopoulos_transformers_2020,sun_retentive_2023} uses the kernel method to replace the exponential kernel $\exp(\boldsymbol{K}_i^\intercal \boldsymbol{Q}_t)$ with a non-negative dot-product $\phi(\boldsymbol{K}_i)^\intercal \phi(\boldsymbol{Q}_t)$. 
\Cref{fig:met:gdkvm} shows that LKVA allows us to express the softmax matching as a linear RNN with a matrix-form hidden state $\mathbf{S}_t$:
\begin{equation}
	\begin{split}		
		\boldsymbol{O}_t    &= \sum_{i=1}^{t} \frac{ \phi(\boldsymbol{K}_i)^{\intercal} \phi(\boldsymbol{Q}_t)}{\sum_{j=1}^{t} \phi(\boldsymbol{K}_j)^{\intercal} \phi(\boldsymbol{Q}_t)} \boldsymbol{V}_i \\
							&= \frac{ (\sum_{i=1}^{t} \boldsymbol{V}_i \phi(\boldsymbol{K}_i)^{\intercal}) \phi(\boldsymbol{Q}_t)}{ (\sum_{j=1}^{t} \phi(\boldsymbol{K}_j)^{\intercal}) \phi(\boldsymbol{Q}_t)}  \\
							&= \frac{ \mathbf{S}_t \phi(\boldsymbol{Q}_t)}{ \boldsymbol{Z}_t^{\intercal} \phi(\boldsymbol{Q}_t)} ,
	\end{split}
	\label{eq:the_linear_attention}
\end{equation}
where $\mathbf{S}_t = \sum_{i=1}^t \boldsymbol{V}_i \phi(\boldsymbol{K}_i)^\intercal \in \mathbb{R}^{C_v \times C_k}$ can be regarded as a two-dimensional recurrent hidden state of fixed size,
and $\boldsymbol{Z}_t=\sum_{j=1}^t\phi(\boldsymbol{K}_j)\in\mathbb{R}^{C_k}$. 
LKVA maps the key space $\mathbb{R}^{C_k}$ and value space $\mathbb{R}^{C_v}$ of each timestep into a more expressive state space $\mathbb{R}^{C_v \times C_k}$.
Some recent linear RNN models \cite{yang2024parallelizing, qin_devil_2022, schlag_linear_2021, parallel-martin} have further provided a simplified linear Transformer formulation:
\begin{equation}
    \begin{split}
	\mathbf{S}_t    	&= \mathbf{S}_{t-1} + \boldsymbol{V}_t \boldsymbol{K}_t^{\intercal} \in \mathbb{R}^{C_v \times C_k}, \\
	\boldsymbol{O}_t	&= \mathbf{S}_t \boldsymbol{K}_t \in \mathbb{R}^{HW \times C_v}.
    \end{split}
    \label{eq:the_recurrent_form}
\end{equation}
Because this recurrent form of linear attention only needs to accumulate and update $\mathbf{S}_t$ along with the associated key-value representations, its computational complexity drops to $\mathcal{O}(t C_v C_k)$. 

The LKVA method possesses linear computational complexity, a feature of critical importance for clinical applications. This ensures the speed and scalability required for processing long-duration echocardiography videos, making real-time analysis possible without sacrificing the model's ability to capture long-range dependencies. The sequential processing approach of LKVA, combined with its highly expressive state representation, is exceptionally well-suited for capturing the continuous and often periodic motion of cardiac structures by progressively accumulating contextual information frame by frame. 
            \begin{figure}[t]
    \centering
        \includegraphics[width=1.0\linewidth]{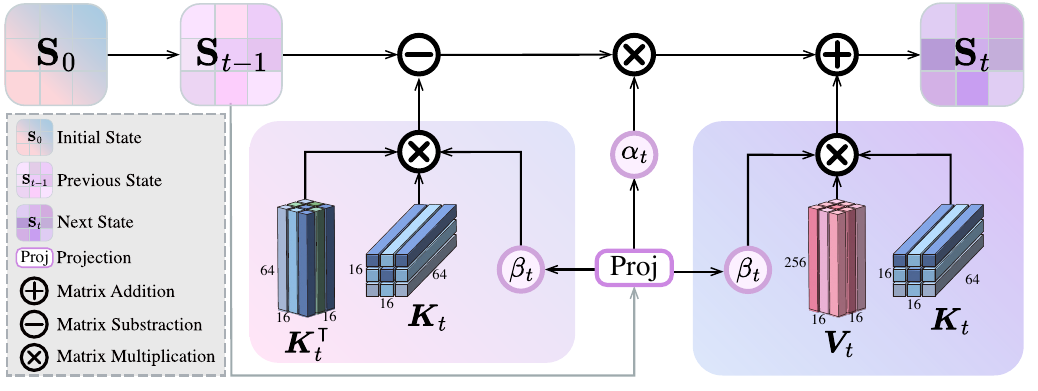}
    \caption{An illustration of Gated Delta Rule (GDR). It leverages a projected $\boldsymbol{\beta}_t$ for efficient key–value associations and frame updates, while a projected $\boldsymbol{\alpha}_t$ quickly adapts to drastic heart shape changes and preserves crucial long-term information.}
    \label{fig:Gated_Delta_Rule}
\end{figure}
    \subsection{Gated Delta Rule}
    \label{subsec:gated_delta_rule}
\paragraph{Delta rule.}
Pure linear attention superimposes all historical information with equal weight. When a sufficient number of tokens are superimposed, the proportion of information from each individual token becomes extremely small. Relying solely on a fixed-size state matrix $\mathbf{S}_{t}$ makes it impossible to accurately reconstruct even an arbitrary output $\boldsymbol{O}_{t}$, causing the state representations to become indistinct and blurred \cite{yang2024parallelizing,schlag_linear_2021,sun_retentive_2023}. 
To mitigate this problem, the model should discard less important associations based on how newly arriving keys interact with existing contents.

In video tasks, from the perspective of key-value retrieval, we first remove the old association for the current frame's key $\boldsymbol{K}_t$ from the previous time-step’s state $\mathbf{S}_{t-1}$, which is represented as $\boldsymbol{V}_{t}^{\text{old}}= \mathbf{S}_{t-1} \boldsymbol{K}_t$. A new value $\boldsymbol{V}_{t}^{\text{new}}$ is obtained by interpolating between the old value and the current frame’s target $\boldsymbol{V}_{t}$, replacing the old value in the state: 
\begin{equation}
	\begin{aligned}
		\mathbf{S}_{t}  &= \mathbf{S}_{t-1} \\
							&\quad  - \underbrace{ \left( \mathbf{S}_{t-1} \boldsymbol{K}_t \right) }_{\boldsymbol{V}_t^{\text{old}}} \boldsymbol{K}_t^{\intercal} 
									+ \underbrace{ \left( \boldsymbol{\beta}_t \boldsymbol{V}_t + \left( \mathbf{I} - \boldsymbol{\beta}_t \right) \mathbf{S}_{t-1} \boldsymbol{K}_t \right)}_{\boldsymbol{V}_t^{\text{new}}} \boldsymbol{K}_t^{\intercal} \\
						&= \mathbf{S}_{t-1} \left( \mathbf{I} - \boldsymbol{\beta}_t \boldsymbol{K}_t \boldsymbol{K}_t^{\intercal} \right)  + \boldsymbol{\beta}_t  \boldsymbol{V}_t \boldsymbol{K}_t^{\intercal},
	\end{aligned}
	\label{eq:delta_rule}
\end{equation}
where $\boldsymbol{\beta}_t \in \mathbb{R}^{C_v \times C_k}$ is a data-dependent matrix, projected from the previous state $\mathbf{S}_{t-1}$, and it represents the soft writing strength. 
When rapid cardiac structure motion or clear boundaries appear in the video, the model is assigned a larger $\boldsymbol{\beta}_t$ value to strengthen the learning of key information in the current frame. Conversely, when the image quality is poor or when images are filled with speckle noise, the model can use it to erase interfering information from the previous frame caused by probe movement or artifacts.
\paragraph{Data-dependent decay.}
In echocardiography video segmentation, the cardiac structures undergo rapid and dramatic dynamic changes. As a result, the model's memory state can easily become saturated with a large volume of complex features, which in turn leads to difficulties in key-value retrieval \cite{yang2025gated,dao2024transformersssmsgeneralizedmodels}.
We project a data-dependent decay matrix $\boldsymbol{\alpha}_t \in \mathbb{R}^{C_v \times C_k}$ from the previous state to control the attenuation of past memory: 
\begin{equation}
    \begin{aligned}
		\mathbf{S}_{t} 	&= 	  \mathbf{S}_{t-1} \left( \boldsymbol{\alpha}_t \left( \mathbf{I} - \boldsymbol{\beta}_t \boldsymbol{K}_t \boldsymbol{K}_t^{\intercal} \right) \right) + \boldsymbol{\beta}_t  \boldsymbol{V}_t \boldsymbol{K}_t^{\intercal}. 
    \end{aligned}
    \label{eq:Gated_Delta_Rule}
\end{equation}
\Cref{fig:Gated_Delta_Rule,eq:Gated_Delta_Rule} show that the introduction of $\boldsymbol{\alpha}_t$ weakens the strong regularization constraint between $\mathbf{S}_t$ and $\mathbf{S}_{t-1}$. This allows the model to selectively forget certain past information and thus dynamically allocate memory resources to the most critical cardiac cycle features.
The $\boldsymbol{\alpha}_t$ enables the model to free up sufficient memory space for new key frames or structures, including frames corresponding to abnormal cardiac rhythms, while preventing irrelevant or redundant historical features from occupying memory.

For echocardiography video segmentation tasks with strong temporal correlations, GDR offers a better balance between memory retention and new information integration compared to simple additive updates. It preserves long-term important information while more rapidly adapting to drastic changes in heart shape, thereby improving the accuracy of spatiotemporal feature extraction. 
            \begin{figure}[t]
    \centering
        \includegraphics[width=1.0\linewidth]{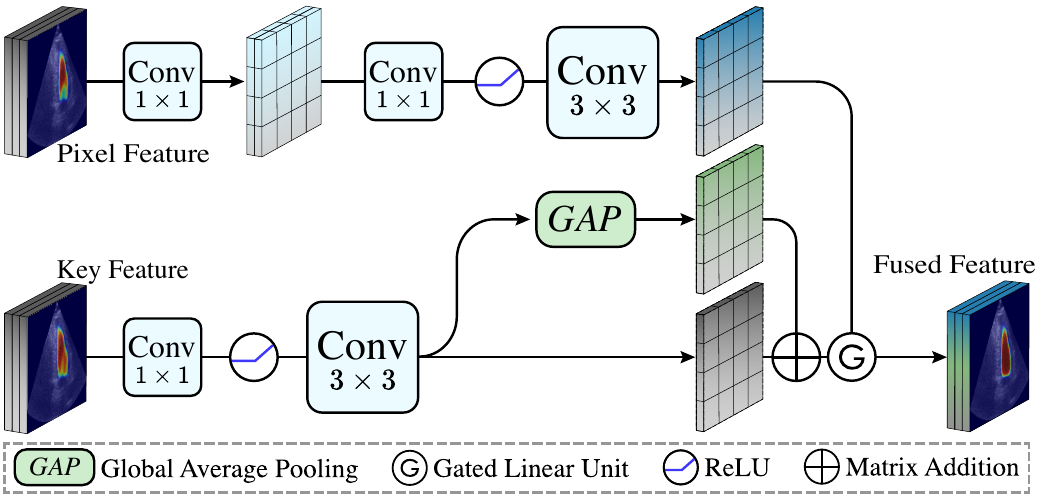}
    \caption{An illustration of Key-Pixel Feature Fusion (KPFF). KPFF fuses the original coarse key features, global key features, and extracted pixel features into key-pixel features, providing enhanced spatiotemporal robustness.}
    \label{fig:kpff}
\end{figure}
    \subsection{Key-Pixel Feature Fusion}
    \label{subsec:kpff}
In echocardiography video segmentation, issues such as speckle noise, artifacts caused by signal attenuation, and low contrast between tissue and blood pose significant challenges to the feature extraction capabilities of models.
Linear models demonstrate rapid and stable memory capabilities in forming key-value associations. However, for inputs with rich spatial structures, such as images, relying solely on this type of associative memory may be insufficient for complex spatial reasoning. Sub-quadratic attention models like xLSTM~\cite{beck:24xlstm} widely utilize small-kernel 1D depth-wise separable convolutions to extract local semantic information. However, these spatial features often suffer from over-smoothing, leading to the loss of some global context.
\Cref{fig:kpff} shows that KPFF merges the convolution outputs learned from local key feature $\mathbf{F}_{\mathrm{K}}$, global key feature $\mathbf{F}_{\mathrm{Global}}$ and pixel-based feature $\mathbf{F}_{\mathrm{Pix}}$, each capturing information at different spatial scales, into a single network branch:
\begin{equation}
    \begin{aligned}
        \mathbf{F}_{\mathrm{Global}}        &= \mathrm{Expand} (\mathrm{GAP} (\mathbf{F}_{\mathrm{K}})), \\
        \mathbf{G}                          &= \sigma(\mathrm{Conv}_{\mathrm{gate}}(\mathbf{F}_{\mathrm{K}} + \mathbf{F}_{\mathrm{Global}})), \\
        \mathbf{F}_{\mathrm{fused}}         &= \mathbf{G} \cdot (\mathbf{F}_{\mathrm{K}} + \mathbf{F}_{\mathrm{Global}}) + (1 - \mathbf{G}) \cdot \mathbf{F}_{\mathrm{Pix}}, 
    \end{aligned}
    \label{eq:kpff}
\end{equation}
where $\sigma(\cdot)$ denotes the sigmoid function. 
When local features $\mathbf{F}_{\mathrm{K}}$ become unreliable due to speckle noise or signal dropout, $\mathbf{F}_{\mathrm{Global}}$ can provide a stable global context as a supplement. By fusing the smoothed global and local features with pixel-level features $\mathbf{F}_{\mathrm{Pix}}$, which retain high-frequency information, KPFF effectively preserves critical diagnostic details and prevents the loss of fine-grained information, such as valve opening and closing or regional wall motion abnormalities. 
            \begin{table*}[!t]
    \centering
    \begin{tabular}{llllllllll}
    \toprule
    \multirow{2}{*}{Method} & \multirow{2}{*}{Venue} & \multicolumn{4}{c}{CAMUS} & \multicolumn{4}{c}{EchoNet-Dynamic} \\ 
    \cmidrule(lr){3-6} \cmidrule(l){7-10} 
     &  & mDice $\uparrow$ & mIoU $\uparrow$ & HD $\downarrow$ & ASD $\downarrow$ & mDice $\uparrow$ & mIoU $\uparrow$ & HD $\downarrow$ & ASD $\downarrow$ \\ 
    \midrule
    XMem++~\cite{bekuzarov2023xmemproductionlevelvideosegmentation} & ICCV'23 & 89.38 & 85.81 & 4.03 & 4.87 & 87.51 & 83.57 & 3.14 & 2.69 \\
    Cutie~\cite{cheng2024puttingobjectvideoobject} & CVPR'24 & 91.09 & 87.97 & 3.89 & 3.74 & 88.96 & 85.63 & 2.89 & 2.24 \\
    VideoMamba~\cite{li2024videomambastatespacemodel} & ECCV'24 & 91.96 & 89.04 & 3.48 & 3.31 & 90.22 & 87.03 & 2.79 & 2.05 \\
    Vision LSTM~\cite{beck:24xlstm} & ICLR'25 & 92.14 & 89.11 & 3.79 & 3.39 & 90.24 & 89.14 & 2.65 & 1.69 \\ 
    \midrule
    PKEchoNet~\cite{wu2023super} & AAAI'23 & 93.49 & 90.95 & 3.42 & 2.93 & 92.60 & 89.89 & 2.53 & 1.48 \\
    DSA~\cite{lin2024dynamic} & TMI'24 & 94.25 & 91.80 & 3.27 & 2.37 & 92.91 & 90.26 & 2.46 & 1.44 \\
    MemSAM~\cite{deng2024memsam} & CVPR'24 & 93.63 & 90.97 & 3.47 & 2.60 & 92.71 & 89.90 & 2.56 & 1.51 \\
    SimLVSeg~\cite{maani2024simlvseg} & UMB'24 & 92.54 & 89.71 & 3.65 & 3.12 & 91.91 & 89.08 & 2.65 & 1.65 \\ 
    \midrule
    \textbf{\mymodel} & - & \textbf{95.11} & \textbf{92.97} & \textbf{3.05} & \textbf{1.98} & \textbf{93.46} & \textbf{90.86} & \textbf{2.38} & \textbf{1.36} \\ 
    \bottomrule
    \end{tabular}
    \caption{Segmentation performance of \mymodel with state-of-the-art methods on CAMUS and EchoNet-Dynamic.}
    \label{tab:cap}
\end{table*}
\section{Experiments}
    \label{sec:experiments}
    \subsection{Setup}
    \label{subsec:setup}
\paragraph{Datasets.}
    \label{para:datasets}
We evaluate our proposed model on two public echocardiography video datasets, CAMUS~\cite{leclerc2019deep} and EchoNet-Dynamic~\cite{ouyang2019echonet}. 
\textbf{CAMUS} contains 500 patient cases, each with apical four-chamber and two-chamber view videos. For each video, it provides frame-by-frame annotations spanning from end-diastole ($ED$) to end-systole ($ES$). These annotations include labeled contours of the left ventricular endocardium, left ventricular epicardium, and the left atrium. 
\textbf{EchoNet-Dynamic} contains 10,030 apical four-chamber view videos. It provides annotations for each video only at two key frames: $ED$ and $ES$. 
We used the original data splits for CAMUS and EchoNet-Dynamic, resizing them to resolutions of $256 \times 256$ and $128 \times 128$, respectively. From each cardiac video, we uniformly sampled 10 frames.

\noindent
\textbf{Evaluation metrics.}
    \label{para:eval}
To ensure a fair and comprehensive evaluation, we assessed the model's performance using both standard segmentation metrics and key clinical indices. The geometric accuracy of the segmentation was evaluated with four common metrics: the mean Dice coefficient (mDice), mean Intersection over Union (mIoU), Hausdorff Distance (HD), and Average Surface Distance (ASD). 
To evaluate the model's clinical utility for left ventricular analysis, we estimated the Left Ventricular Ejection Fraction ($LV_{EF}$) from the segmentation masks using Simpson's rule~\cite{LANG20151}. We then assessed the quality of these estimations by calculating the Pearson correlation coefficient (corr), mean bias (bias), and standard deviation (std) between the predicted and ground truth $LV_{EF}$ values.

\noindent
\textbf{Implementation details.}
    \label{para:imp}
All training runs on a single RTX 3090 GPU. 
For CAMUS and EchoNet-Dynamic, we conducted training over 1500 iterations. 
The AdamW optimizer~\cite{loshchilov2019decoupledweightdecayregularization} was used with a learning rate of 1e-4, a batch size of 10 and a variety of data augmentation techniques were applied, including gamma augmentation, random scaling, random rotation, and random contrast adjustment, each with a probability of 0.5. We clip the global gradient norm to $\lambda$ = 3 and use stable data augmentation. 
We use a combined loss function of cross-entropy and soft dice loss with equal weighting following \citet{cheng2024puttingobjectvideoobject}.
    \subsection{Comparison with State-of-the-art Methods}
    \label{subsec:comparisons}
We compare GDKVM with four task-specific methods (PKEchoNet~\cite{wu2023super}, DSA~\cite{lin2024dynamic}, MemSAM~\cite{deng2024memsam}, SimLVSeg~\cite{maani2024simlvseg}) and four related approaches (Xmem++~\cite{bekuzarov2023xmemproductionlevelvideosegmentation}, Cutie~\cite{cheng2024puttingobjectvideoobject}, VideoMamba~\cite{li2024videomambastatespacemodel}, Vision LSTM~\cite{beck:24xlstm}). 

These prior works incorporate memory mechanisms, time-series modeling, or efficient attention modules and have advanced medical video segmentation. However, they still face challenges in capturing complex spatial variations and retrieving temporal information accurately. GDKVM addresses these limitations by introducing key-value associations within a linear matching framework, which reduces the typical high computational overhead of conventional attention. In \cref{tab:cap}, GDKVM delivers higher overall metrics on two datasets and outperforms four widely used task-specific models. In \cref{fig:fps}, GDKVM achieves an mDice of 95.11 at 37 FPS with 35.2~M parameters. GDKVM leverages GDR for dynamic memory management to discard irrelevant information promptly and update essential memory content efficiently, matching or surpassing specialized models in segmentation performance. This design avoids bottlenecks arising from incomplete memory mechanisms or overly complex training procedures, allowing GDKVM to maintain both accuracy and speed.
            \begin{figure}[t]
    \centering
        \includegraphics[width=1.0\linewidth]{fig/fps/fig_fps.pdf}
    \caption{Comparison of different methods in terms of speed (FPS) and mean Dice score on CAMUS, with bubble size representing the number of parameters.}
    \label{fig:fps}
\end{figure}
        
\begin{figure}[thb]
    \centering
        \begin{subfigure}{0.49\linewidth}
		\centering
		\includegraphics[width=1.0\linewidth]{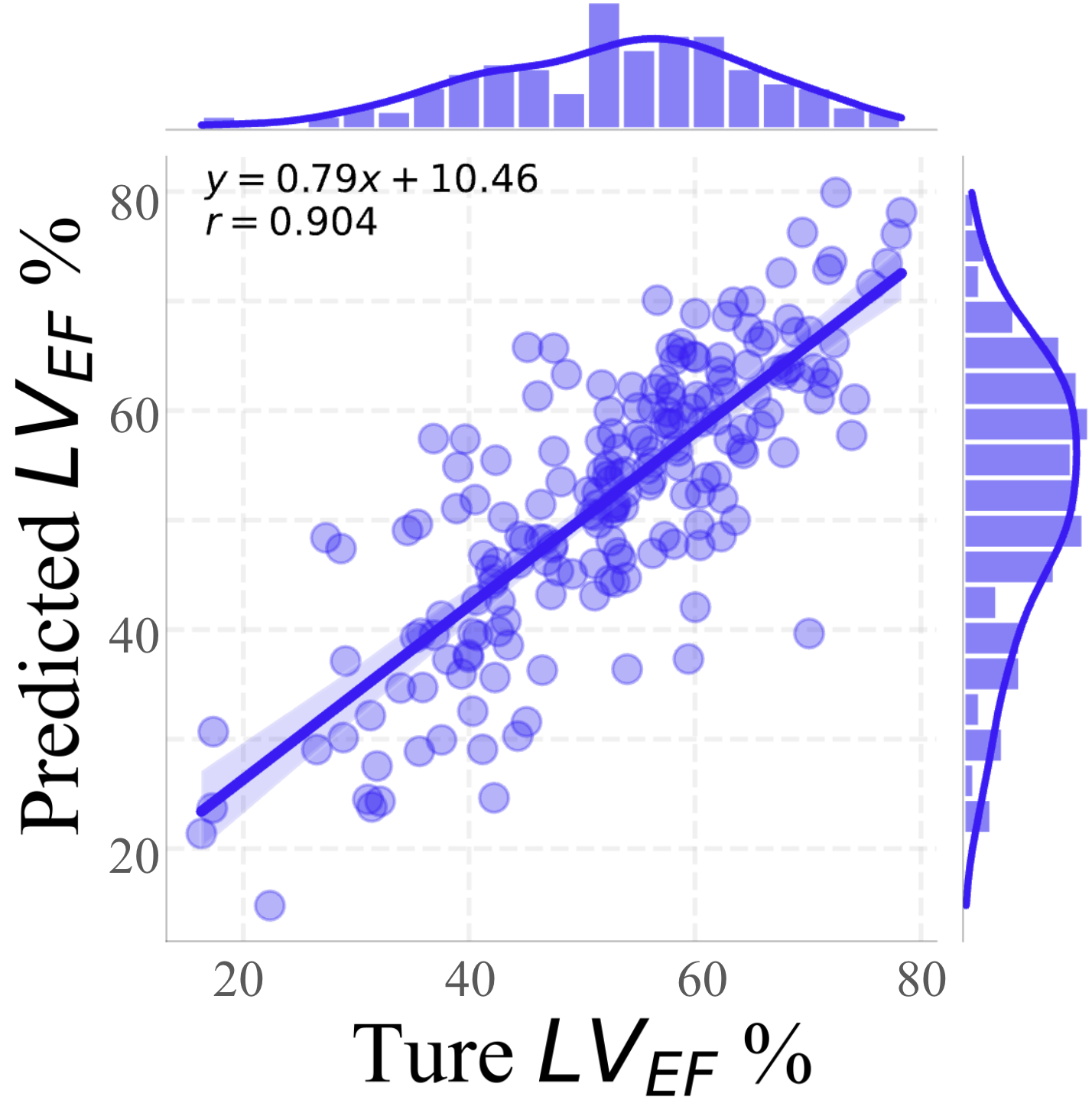}
		\label{fig:linear_regression:camus}
	\end{subfigure}
	\begin{subfigure}{0.49\linewidth}
		\centering
		\includegraphics[width=1.0\linewidth]{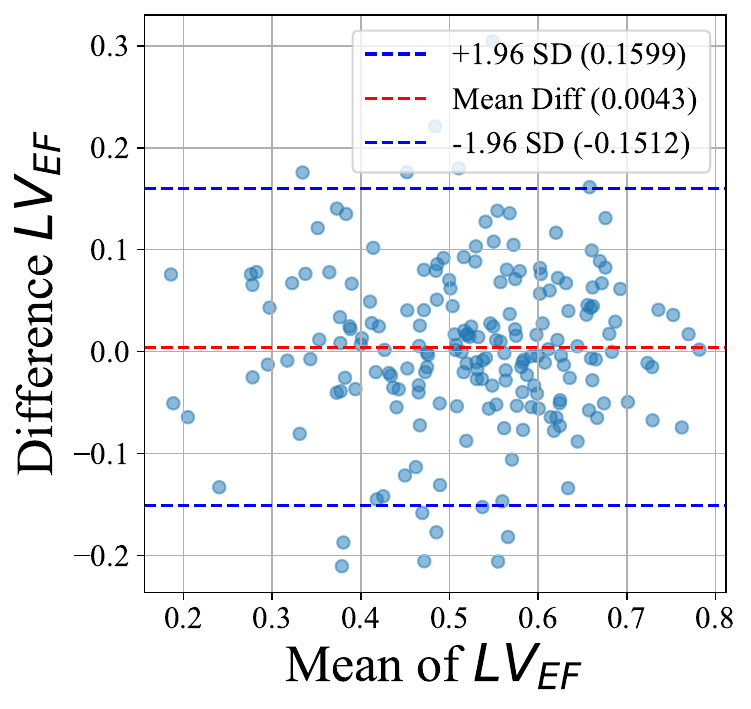}
		\label{fig:ba:camus}
	\end{subfigure}
    \caption{Linear regression and Bland–Altman plots for clinical metric $LV_{EF}$ on CAMUS.}
    \label{fig:LV-BA}
\end{figure}

\noindent
\textbf{Clinical metric.}
Based on the segmentation results, we estimated the clinical metric $LV_{EF}$. \Cref{tab:tab2} shows that GDKVM obtains the highest Pearson correlation coefficient of 0.904 and the lowest bias and standard deviation ($-0.19 \pm 11.3\%$). This indicates a strong agreement between our predictions and the ground truth. \Cref{fig:LV-BA} provides a visual representation, where the linear regression plot shows a strong correlation, and the Bland-Altman plot demonstrates high consistency.
            \begin{table}[t]
    \centering
    \renewcommand{\arraystretch}{0.9} 
    \begin{tabular}{lll}
    \toprule
    \multirow{2}{*}{Method} & \multicolumn{2}{c}{CAMUS} \\ 
    \cmidrule(l){2-3} 
     & corr $\uparrow$ & bias $\pm$ std (\%) \\ 
    \midrule
    XMem++~\cite{bekuzarov2023xmemproductionlevelvideosegmentation} & 0.746 & 1.70 $\pm$ 21.9 \\
    Cutie~\cite{cheng2024puttingobjectvideoobject}                 & 0.787 & 1.67 $\pm$ 21.7 \\
    VideoMamba~\cite{li2024videomambastatespacemodel}               & 0.780 & -4.49 $\pm$ 19.4 \\
    Vision LSTM~\cite{beck:24xlstm}                                 & 0.806 & -0.31 $\pm$ 18.8 \\ 
    \midrule
    PKEchoNet~\cite{wu2023super}                                    & 0.862 & -1.53 $\pm$ 16.4 \\
    DSA~\cite{lin2024dynamic}                                       & 0.891 & 0.86 $\pm$ 13.4 \\
    MemSAM~\cite{deng2024memsam}                                    & 0.878 & -0.89 $\pm$ 12.3 \\
    SimLVSeg~\cite{maani2024simlvseg}                               & 0.895 & 1.83 $\pm$ 13.8 \\ 
    \midrule
    \textbf{\mymodel}                                                  & \textbf{0.904} & \textbf{-0.19 $\pm$ 11.3} \\ 
    \bottomrule
    \end{tabular}
    \caption{Clinical metrics comparison against different state-of-the-art methods on CAMUS.}
    \label{tab:tab2}
\end{table}
            \begin{figure*}[!t]
    \centering
    \includegraphics[width=1.0\textwidth]{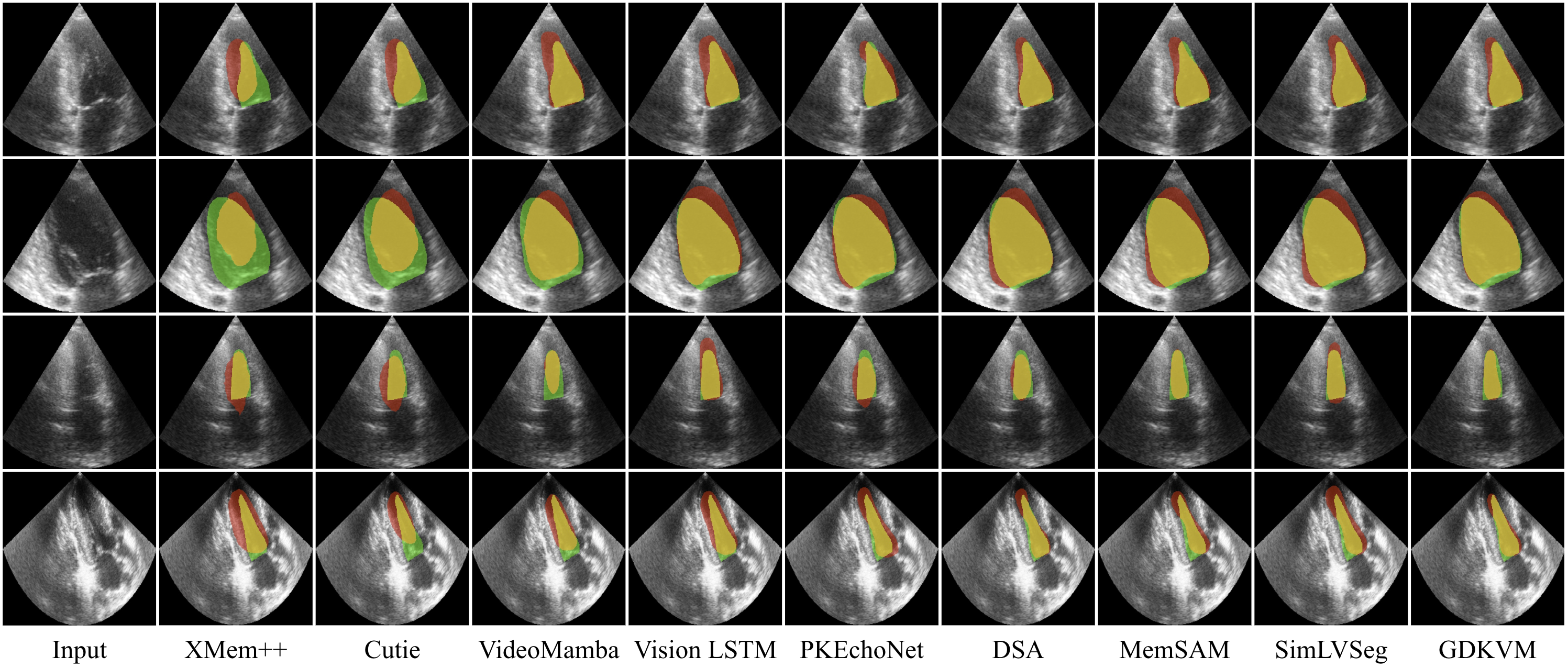}
    \caption{Visual comparison with state-of-the-art methods on the CAMUS test set. Green, red, and yellow regions represent the ground truth, prediction, and overlapping regions, respectively.}
    \label{fig:vis_cap}
\end{figure*}
        
\noindent
\textbf{Visual comparison with SOTA.}
A visual comparison of GDKVM with other state-of-the-art methods is presented in \cref{fig:vis_cap}, illustrating the ability of GDKVM to preserve structural integrity and delineate precise boundaries in difficult frames. Although prior approaches have significantly advanced heart cavity segmentation, they can still produce over-segmentation (including extraneous areas beyond the heart cavity) or under-segmentation (missing parts of the ventricle). In these challenging cases, existing methods sometimes deviate from the ground truth shape by forming gaps or protrusions. GDKVM, by contrast, adheres more closely to the anatomical contours, generating cleaner masks with noticeably fewer artifacts. This property yields higher segmentation fidelity on frames where earlier approaches often falter.
            \begin{figure}[t]
    \centering
    \includegraphics[width=0.99\linewidth]{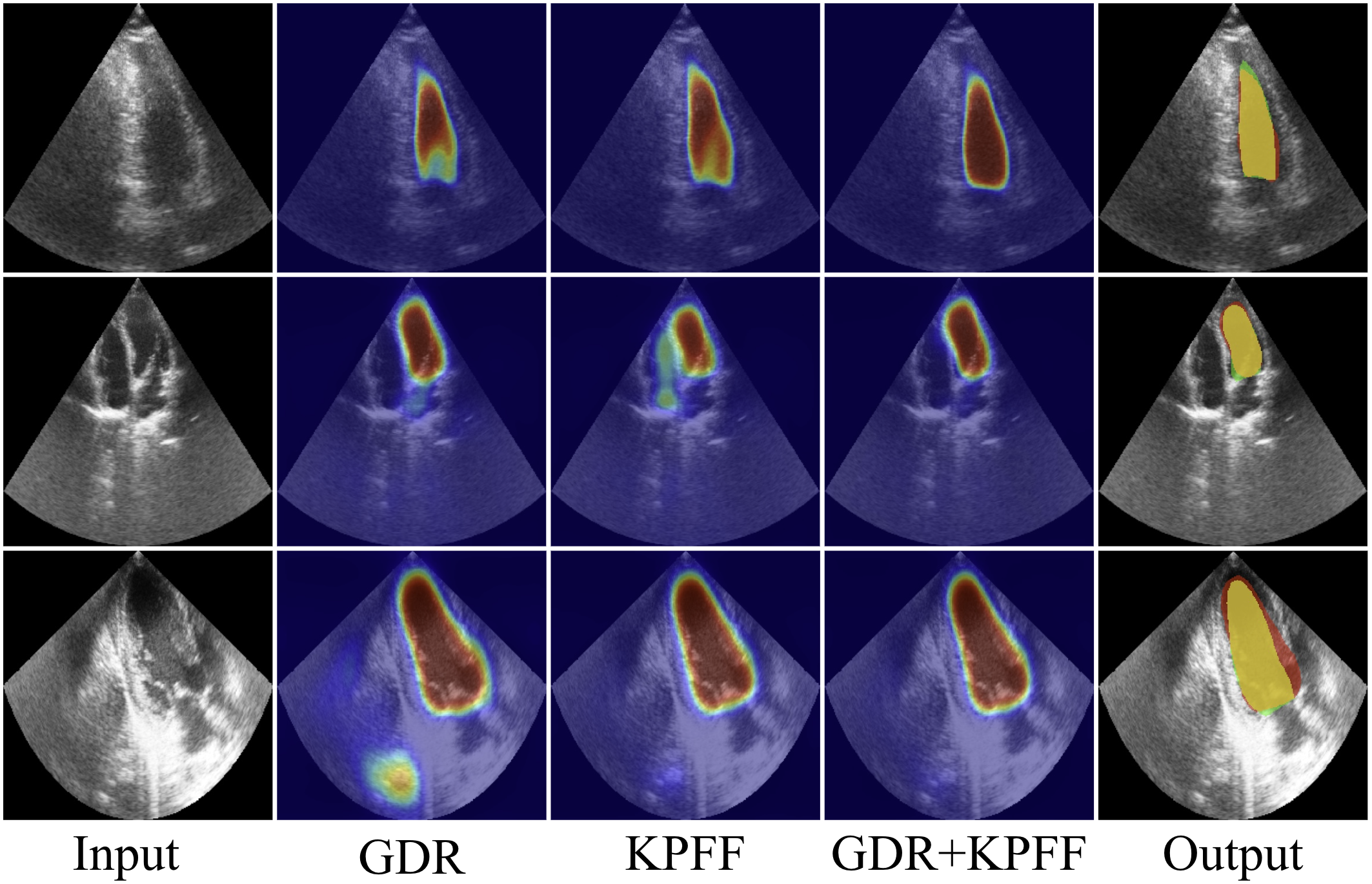}
    \caption{Visualization of segmentation results using different ablation strategies of \mymodel on CAMUS. The heatmap colors indicate prediction confidence, with warmer colors denoting higher confidence.}
    \label{fig:abl}
\end{figure}
            \begin{table}[t]
    \centering
    \footnotesize
    \setlength{\tabcolsep}{2.5mm}
    \renewcommand\arraystretch{1.1}
    
    \begin{tabular}{ccc|llll}
    \hline
    LKVA & GDR & KPFF & mDice & mIoU & HD & ASD \\ 
    \hline
    \checkmark & & & 93.10 & 90.46 & 3.65 & 2.85 \\
    \checkmark & \checkmark & & 94.49 & 92.11 & 3.21 & 2.19 \\
    \checkmark & & \checkmark & 93.30 & 90.78 & 3.55 & 2.74 \\ 
    \hline
    \checkmark & \checkmark & \checkmark & \textbf{95.11} & \textbf{92.97} & \textbf{3.05} & \textbf{1.98} \\ 
    \hline
    \end{tabular}
    \caption{Ablation study on different components of \mymodel on CAMUS.}
    \label{tab:abl}
\end{table}
    \subsection{Ablation Studies}
    \label{subsec:abl}
To evaluate each component of \mymodel, we performed ablation studies on CAMUS, as shown in \cref{fig:abl,tab:abl}. 
        \paragraph{Effectiveness of Key-Pixel Feature Fusion.}
\begin{sloppypar}
KPFF mechanism is critical for handling the significant scale variations and spatial transformations inherent in echocardiography videos. Disabling this mechanism and relying on a single key feature extraction approach causes a marked degradation in segmentation performance (\cref{tab:abl}). The model becomes particularly vulnerable to image noise and distortion, leading to less accurate boundary delineation, as visually evidenced in \cref{fig:abl}. These results establish that the proposed feature fusion is essential for achieving robust and precise segmentation in such dynamic clinical data.
\end{sloppypar}
            \begin{table}[t]
\renewcommand{\arraystretch}{1.4}
\resizebox{\columnwidth}{!}{%
\begin{tabular}{@{}llll@{}}
\hline
\textbf{Strategy} & \textbf{Recurrence Equation} & \textbf{mDice} & \textbf{Inf. Time} \\ \hline
Baseline & $\mathbf{S}_{t} = \mathbf{S}_{t-1} + \boldsymbol{V}_t \boldsymbol{K}_t^{\intercal}$ & 93.30 & 151.61 \\
Sanity Check & $\mathbf{S}_{t} = \mathbf{S}_{t-1} - (\mathbf{S}_{t-1} \boldsymbol{K}_t)\boldsymbol{K}_t^{\intercal} + \boldsymbol{V}_t \boldsymbol{K}_t^{\intercal}$ & 74.68 & 155.09 \\ \hline
w/o $\alpha_t$ & $\mathbf{S}_{t} = \mathbf{S}_{t-1} \left( \mathbf{I} - \beta_t \boldsymbol{K}_t \boldsymbol{K}_t^{\intercal} \right) + \beta_t \boldsymbol{V}_t \boldsymbol{K}_t^{\intercal}$ & 94.57 & 158.77 \\
w/o $\beta_t$ & $\mathbf{S}_{t} = \alpha_t \mathbf{S}_{t-1} + \boldsymbol{V}_t \boldsymbol{K}_t^{\intercal}$ & 94.26 & 156.90 \\
GDR & $\mathbf{S}_{t} = \mathbf{S}_{t-1} \left( \alpha_t \left( \mathbf{I} - \beta_t \boldsymbol{K}_t \boldsymbol{K}_t^{\intercal} \right) \right) + \beta_t \boldsymbol{V}_t \boldsymbol{K}_t^{\intercal}$ & 95.11 & 160.62 \\ \hline
\end{tabular}%
}
    \caption{Effectiveness of different state update strategies. We report the segmentation performance (mDice) and inference time in milliseconds (ms).}
    \label{tab:strategy}
\end{table}
        \paragraph{Effectiveness of Gated Delta Rule.}
In \cref{tab:strategy}, we compare several linear update strategies for the memory state. 
Directly replacing old memory with new information yields a Dice score of 74.68, indicating that a simple structural change in the update formula is insufficient. Introducing either $\boldsymbol{\alpha}_t$ or $\boldsymbol{\beta}_t$ alone slightly improves performance, suggesting that selectively forgetting old content or reinforcing new features helps capture dynamic cardiac boundaries. Combining $\boldsymbol{\alpha}_t$ and $\boldsymbol{\beta}_t$ further provides bidirectional control: $\boldsymbol{\alpha}_t$ governs the decay of stored shapes to remove noise or errors, and $\boldsymbol{\beta}_t$ sets the strength of newly written features to accommodate abrupt structural changes. 
As a result, GDR reaches a Dice score of 95.11, surpassing the baseline by about 1.7 percentage points.

\begin{figure}[t]
    \centering
        \begin{subfigure}{0.49\linewidth}
		\centering
		\includegraphics[width=1.0\linewidth]{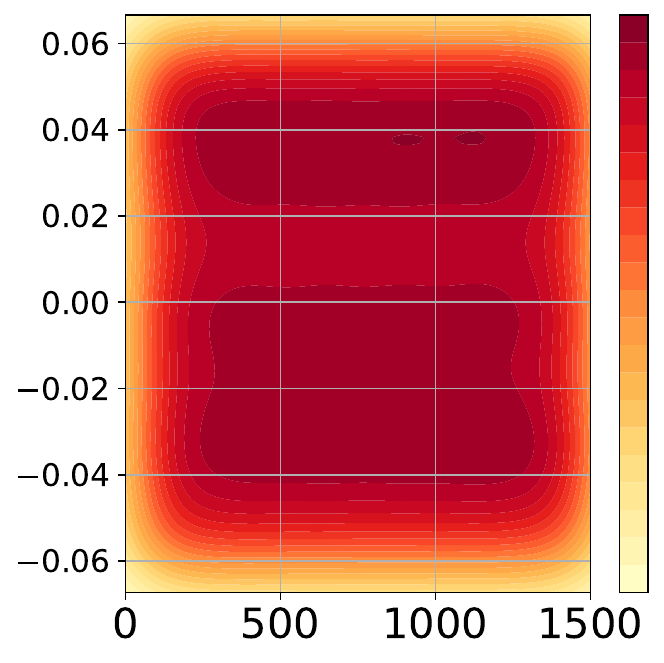}
		\caption{Distribution of parameter $\boldsymbol{\alpha}_t$. Darker regions indicate higher density.} 
		\label{fig:wei:para_a}
	\end{subfigure}
	\begin{subfigure}{0.49\linewidth}
		\centering
		\includegraphics[width=1.0\linewidth]{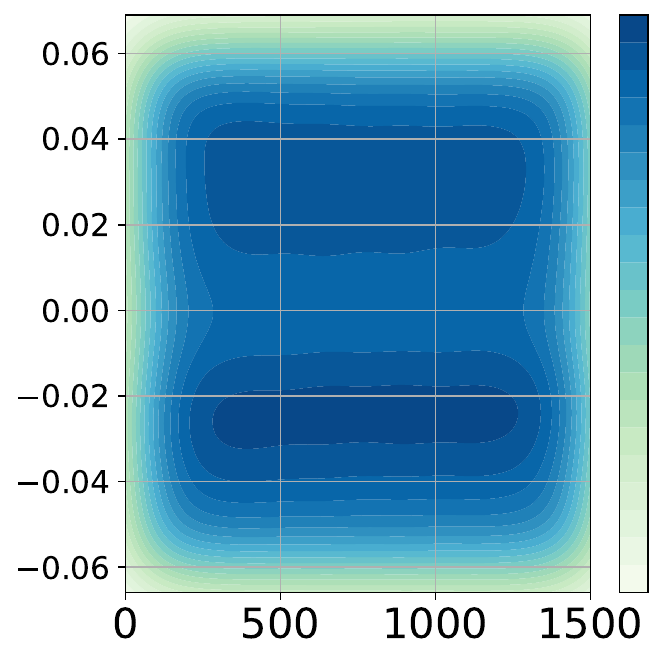}
		\caption{Distribution of parameter $\boldsymbol{\beta}_t$. Darker regions indicate higher density.} 
		\label{fig:wei:para_b}
	\end{subfigure}
	\begin{subfigure}{0.49\linewidth}
		\centering
		\includegraphics[width=1.0\linewidth]{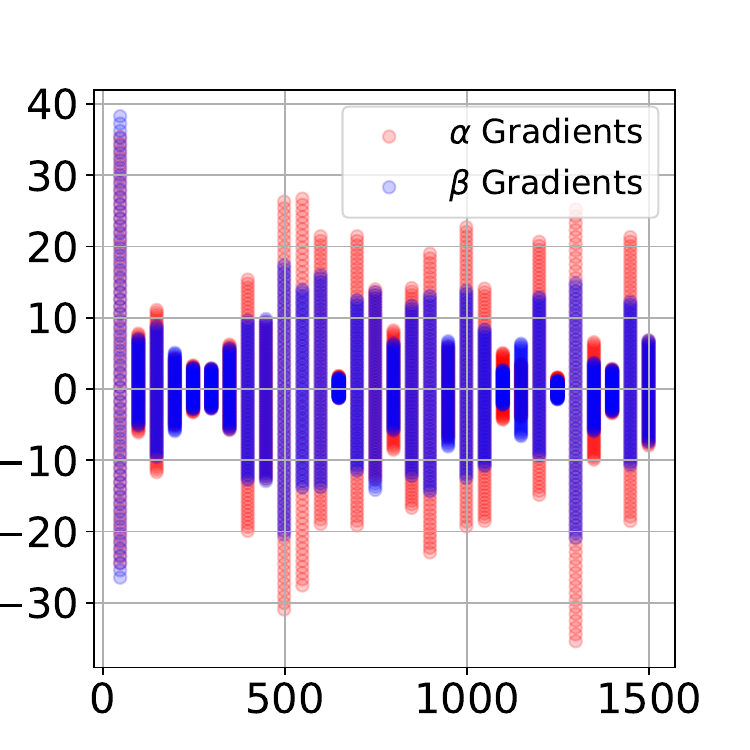}
		\caption{Distribution of gradient $\boldsymbol{\alpha}_t$ and $\boldsymbol{\beta}_t$.}
		\label{fig:wei:Grad}
	\end{subfigure}
        \hspace{0.03mm}
	\begin{subfigure}{0.49\linewidth}
		\centering
		\includegraphics[width=1.0\linewidth]{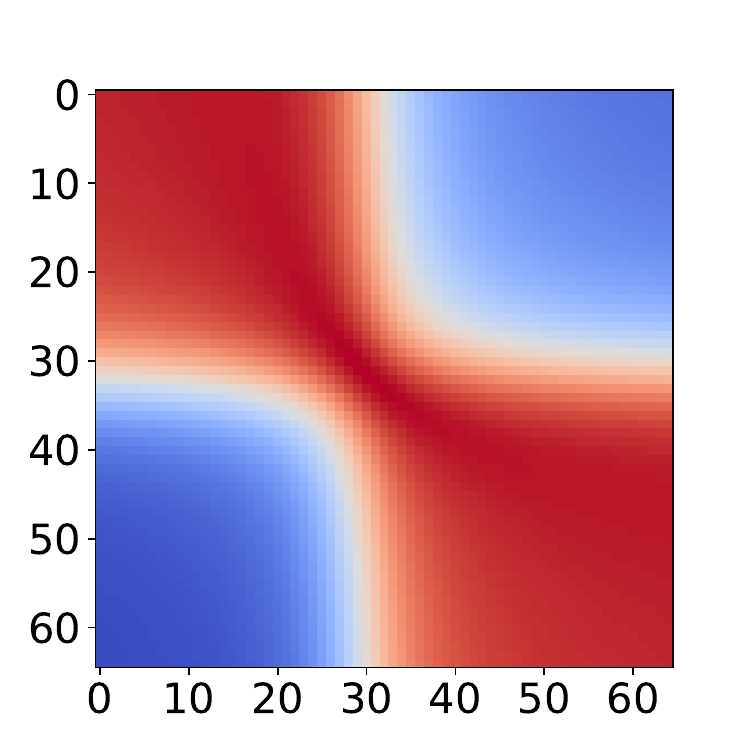}
		\caption{Pearson Correlation of gradient $\boldsymbol{\alpha}_t$ and $\boldsymbol{\beta}_t$. }
		\label{fig:wei:pearson}
	\end{subfigure}
    \caption{Weights of parameters $\boldsymbol{\alpha}_t$ and $\boldsymbol{\beta}_t$ over training steps on CAMUS.}
    \label{fig:wei}
\end{figure}

In \cref{fig:wei:para_a,fig:wei:para_b}, $\boldsymbol{\alpha}_t$ and $\boldsymbol{\beta}_t$ concentrate at specific intervals. During stable frames, both decay and writing intensities remain moderate. In frames with blurred boundaries or heavy noise, $\boldsymbol{\alpha}_t$ increases forgetting to discard flawed information. In cases of sudden contour changes, $\boldsymbol{\beta}_t$ boosts new feature updates. Large gradient jumps, such as +40 or -30 in \cref{fig:wei:Grad}, illustrate the importance of robust gating. In \cref{fig:wei:pearson}, the correlation pattern forms block-like and diagonal structures, reflecting the coordinated adjustment of $\boldsymbol{\alpha}_t$ and $\boldsymbol{\beta}_t$. This mechanism refines echocardiographic boundaries, filters noise during complex cardiac motion, and retains useful features in stable periods for more precise segmentation.
\subsection{Discussions and Limitations}
    \label{subsec:discussion}
\begin{figure}[t]
    \centering
        \includegraphics[width=0.99\linewidth]{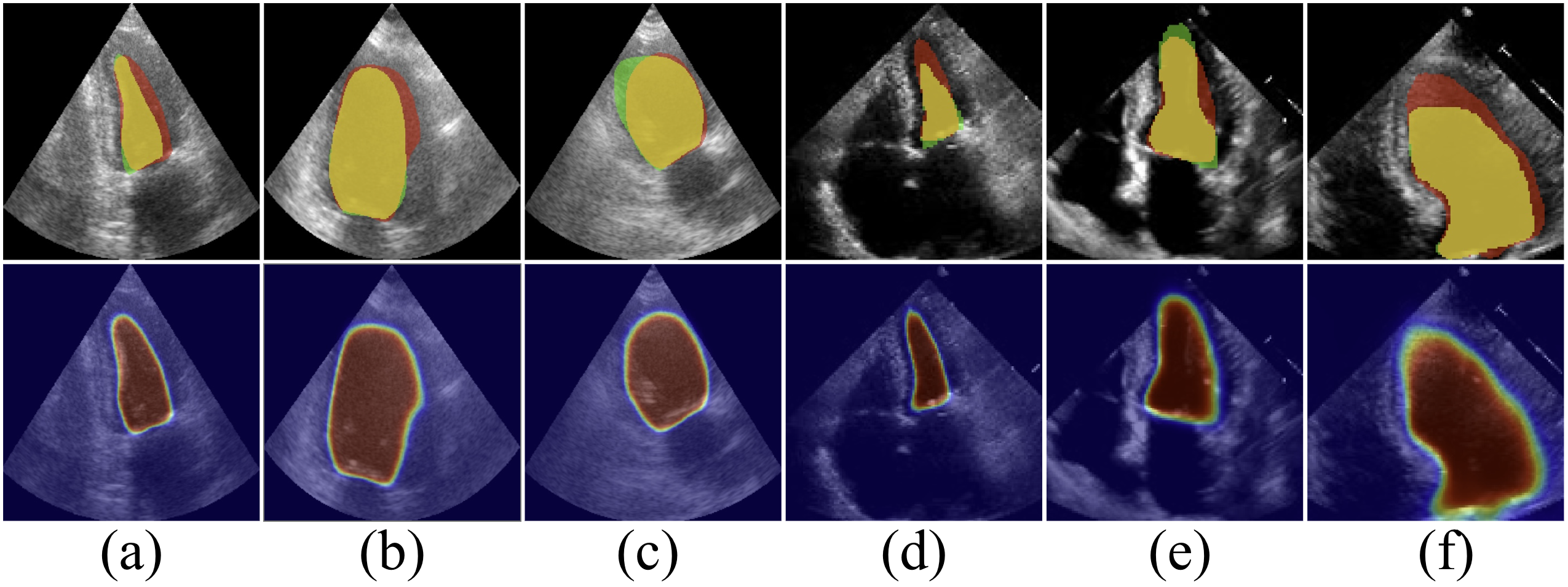}
    \caption{Failure cases on CAMUS (a-c) and EchoNet-Dynamic (d-f) test sets.}
    \label{fig:fail}
\end{figure}

Our work has some limitations. 
While the model performs well in ultrasound cardiac video segmentation, its full potential in other medical video object segmentation scenarios remains to be explored. This includes testing on fully-supervised and semi-supervised benchmarks, which would allow us to evaluate the model's performance in accurately modeling long-term temporal sequences. Future work could also focus on developing metrics that better assess the temporal consistency of sequences.
\Cref{fig:fail} shows that the model sometimes over-relies on its own predictions during the final optimization stage. For some difficult samples, this hinders the precise delineation of contours. A potential direction is to develop more flexible boundary definitions or adaptive corrections to address these artifacts.
Our current intermediate state matrix is not a standard square matrix, which restricts chunkwise parallelization and hardware acceleration. Future work could explore a square matrix design to enable more efficient parallel operators, reduce training time, and improve inference speed.
\section{Conclusion}
    \label{sec:conclusion}
This work describes \mymodel, a novel linear key-value memory network designed for efficient and robust echocardiography video segmentation. 
\mymodel directly addresses the clinical need for real-time analysis by leveraging a Linear Key-Value Association (LKVA) to achieve linear computational complexity. To accurately model cardiac dynamics, Gated Delta Rule (GDR) employs two data-dependent decays to intelligently update and forget memory, capturing rapid motion while maintaining temporal consistency. Key-Pixel Feature Fusion (KPFF) module enhances robustness against image artifacts and noise by merging features from different receptive fields.
Experiments demonstrate that \mymodel consistently achieves state-of-the-art segmentation accuracy while remaining highly efficient across two echocardiography video datasets (CAMUS and EchoNet-Dynamic).
\section*{Acknowledgements}
This work was supported partly by the National Natural Science Foundation of China (No. 62273241), the Natural Science Foundation of Guangdong Province, China (No. 2024A1515011946), and the Hong Kong RGC Collaborative Research Fund (project no. C5055-24G). 

{
    \small
    \bibliographystyle{ieeenat_fullname}
    \bibliography{main}
}

\clearpage
\appendix
\maketitlesupplementary
\section{Ejection Fraction Estimation}
The Left Ventricular Ejection Fraction ($LV_{EF}$) is a critical measure of cardiac pump function. Its calculation requires determining the End-Diastolic Volume ($V_{ED}$), the volume of the left ventricle when it is maximally filled, and the End-Systolic Volume ($V_{ES}$), the volume when it is maximally contracted.

The general formula to compute the $LV_{EF}$ is:
\begin{equation}
   LV_{EF} = \frac{V_{ED} - V_{ES}}{V_{ED}} \times 100\%.
   \label{eq:lvef}
\end{equation}
\paragraph{Biplane Simpson's method}
This method is the gold standard recommended by the American Society of Echocardiography~\cite{LANG20151} because it involves fewer geometric assumptions and is more accurate, especially for abnormally shaped ventricles.
The method uses two orthogonal views: the apical four-chamber (a4c) and the apical two-chamber (a2c) view. The left ventricle is divided into a series of $n$ disks of equal height along its long axis, $L$.
Each disk is assumed to be an elliptical cylinder. The total volume $V$ is the sum of the volumes of all disks. The volume is calculated using the diameters of the $i$-th disk measured from the four-chamber ($D_i^{4c}$) and two-chamber ($D_i^{2c}$) views.
The volume formula is given by:
\begin{equation}
  V = \frac{\pi}{4} \sum_{i=1}^{n} \left( D_i^{4c} \cdot D_i^{2c} \cdot \frac{L}{n} \right).
  \label{eq:biplane}
\end{equation}
\paragraph{Single-plane Simpson's method}
This is a simplified version of the method, applied when only one view (typically the apical four-chamber view) is of sufficient quality for analysis.
This method uses only a single view and assumes the left ventricle is a solid of revolution. Therefore, each disk is treated as a perfect circular cylinder.
The volume calculation relies on the diameter of each disk ($D_i^{4c}$) as measured from the single available plane.
The volume formula is:
\begin{equation}
  V = \frac{\pi}{4} \sum_{i=1}^{n} \left( (D_i^{4c})^2 \cdot \frac{L}{n} \right).
  \label{eq:singleplane}
\end{equation}

For both methods, $V_{ED}$ and $V_{ES}$ are calculated separately using the appropriate formula. These values are then used in the general $LV_{EF}$ \cref{eq:lvef} to determine the ejection fraction.
\section{Experiments Continued}
        \subsection{The Clinical Metric}
\begin{figure}[hptb]
    \centering
        \begin{subfigure}{0.49\linewidth}
		\centering
		\includegraphics[width=1.0\linewidth]{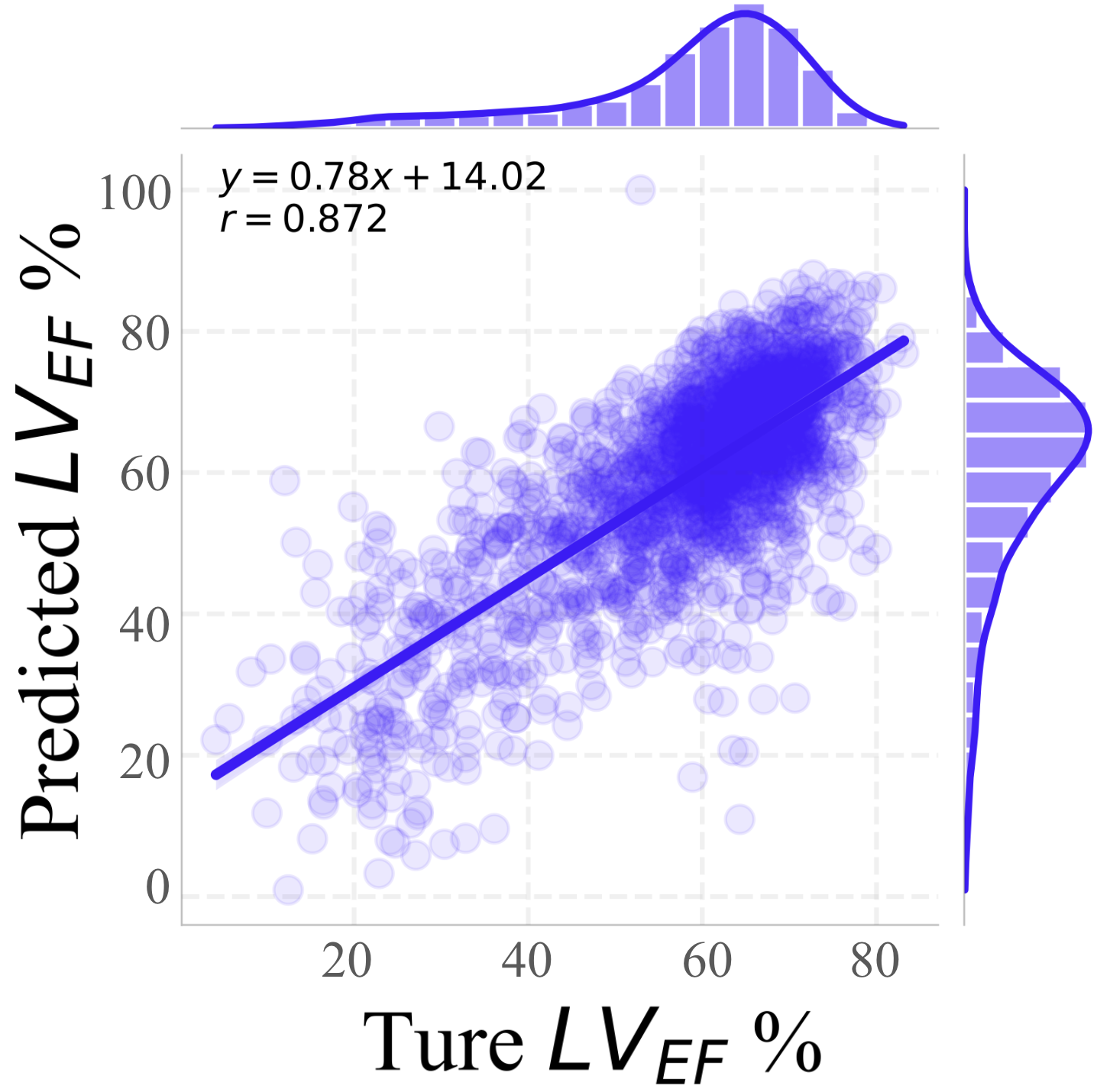}
		\label{fig:linear_regression:echo}
	\end{subfigure}
	\begin{subfigure}{0.49\linewidth}
		\centering
		\includegraphics[width=1.0\linewidth]{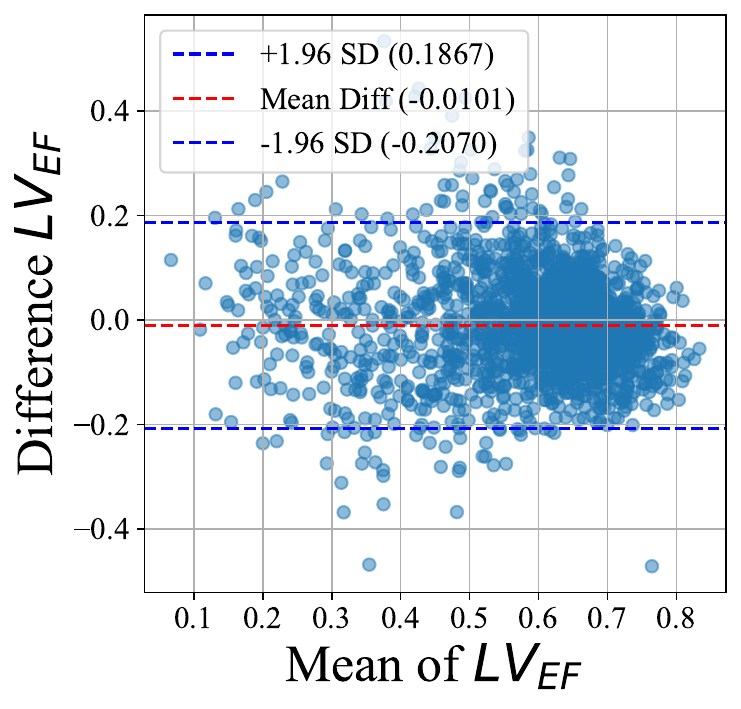}
		\label{fig:ba:echo}
	\end{subfigure}
    \caption{Linear regression and Bland-Altman plots for clinical metric $LV_{EF}$ on EchoNet-Dynamic.}
    \label{fig:LV-BA:echo}
\end{figure}

We also performed linear regression and Bland-Altman plots for the clinical metric $LV_{EF}$ on EchoNet-Dynamic.
\Cref{tab:clinical:echo,fig:LV-BA:echo} show that our method demonstrated high agreement and favorable clinical metrics.
                \begin{table}[hptb]
    \centering
    \begin{tabular}{lll}
    \toprule
    \multirow{2}{*}{Method} & \multicolumn{2}{c}{EchoNet-Dynamic} \\ 
    \cmidrule(l){2-3} 
     & corr $\uparrow$ & bias $\pm$ std (\%) \\ 
    \midrule
    XMem++~\cite{bekuzarov2023xmemproductionlevelvideosegmentation}~\textcolor{gray}{\scriptsize ICCV'23} & 0.692 & 5.77 $\pm$ 10.4 \\
    Cutie~\cite{cheng2024puttingobjectvideoobject}~\textcolor{gray}{\scriptsize CVPR'24}               & 0.695 & 4.30 $\pm$ 11.7 \\
    VideoMamba~\cite{li2024videomambastatespacemodel}~\textcolor{gray}{\scriptsize ECCV'24}         & 0.764 & -2.92 $\pm$ 9.38 \\
    Vision LSTM~\cite{beck:24xlstm}~\textcolor{gray}{\scriptsize ICLR'25}                          & 0.768 & -2.42 $\pm$ 9.10 \\ 
    \midrule
    PKEchoNet~\cite{wu2023super}~\textcolor{gray}{\scriptsize AAAI'23}                             & 0.852 & 1.29 $\pm$ 9.62 \\
    DSA~\cite{lin2024dynamic}~\textcolor{gray}{\scriptsize TMI'24}                                 & 0.868 & 1.42 $\pm$ 9.18 \\
    MemSAM~\cite{deng2024memsam}~\textcolor{gray}{\scriptsize CVPR'24}                             & 0.859 & 1.09 $\pm$ 9.44 \\
    SimLVSeg~\cite{maani2024simlvseg}~\textcolor{gray}{\scriptsize UMB'24}                         & 0.794 & 0.91 $\pm$ 9.56 \\ 
    \midrule
    \textbf{\mymodel}                                                                                 & \textbf{0.872} & \textbf{-0.70 $\pm$ 9.15} \\ 
    \bottomrule
    \end{tabular}
    \caption{Clinical metrics comparison against different state-of-the-art methods on EchoNet-Dynamic.}
    \label{tab:clinical:echo}
\end{table}
        \subsection{Visual Comparison with SOTA}
\begin{figure*}[!t]
    \centering
    \includegraphics[width=1.0\textwidth]{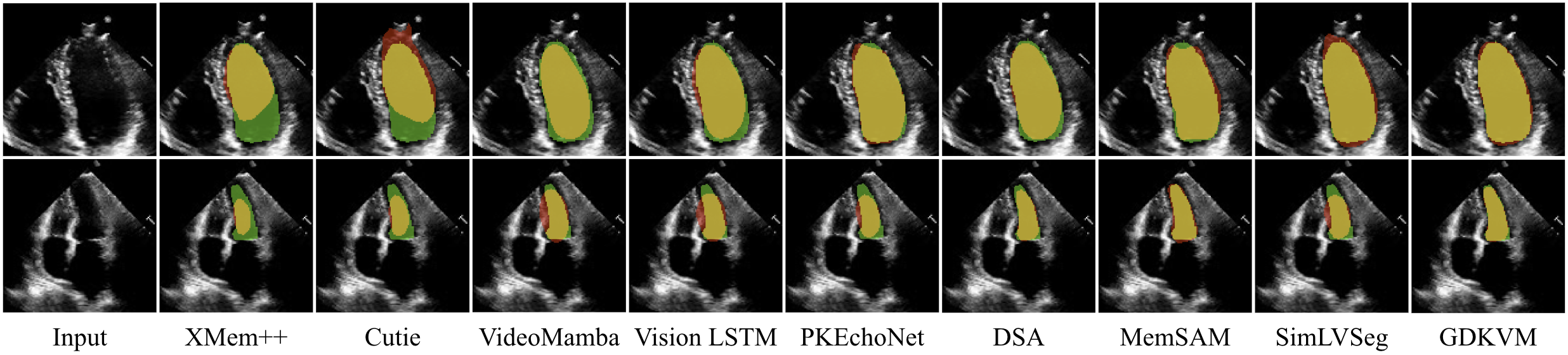}
    \caption{Visual comparison with state-of-the-art methods on the EchoNet-Dynamic test set. Green, red, and yellow regions represent the ground truth, prediction, and overlapping regions, respectively.}
    \label{fig:cap:echo}
\end{figure*}

We also conducted experiments on EchoNet-Dynamic, and the visualization is shown in \cref{fig:cap:echo}.
        \subsection{Visualization of the Weights}
\begin{figure}[thpb]
    \centering
        \begin{subfigure}{0.49\linewidth}
		\centering
		\includegraphics[width=1.0\linewidth]{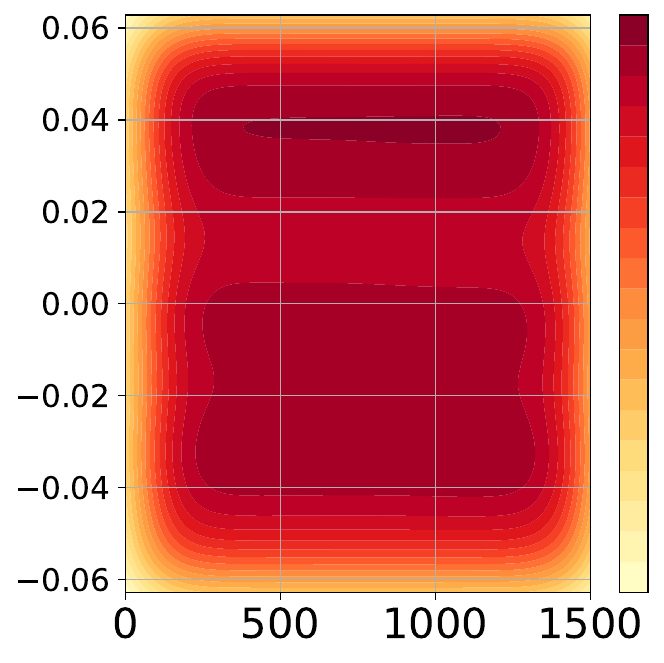}
		\caption{Distribution of parameter $\boldsymbol{\alpha}_t$. Darker regions indicate higher density.} 
		\label{fig:wei:para_a:echo}
	\end{subfigure}
	\begin{subfigure}{0.49\linewidth}
		\centering
		\includegraphics[width=1.0\linewidth]{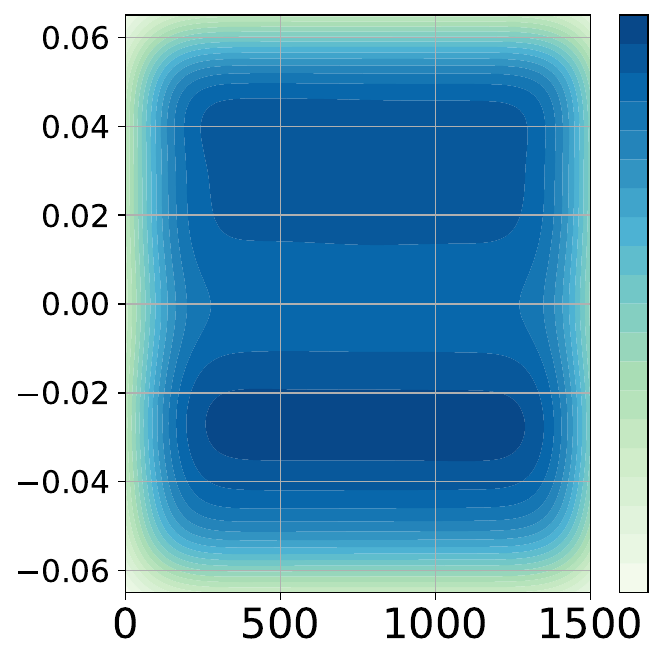}
		\caption{Distribution of parameter $\boldsymbol{\beta}_t$. Darker regions indicate higher density.} 
		\label{fig:wei:para_b:echo}
	\end{subfigure}
	\begin{subfigure}{0.49\linewidth}
		\centering
		\includegraphics[width=1.0\linewidth]{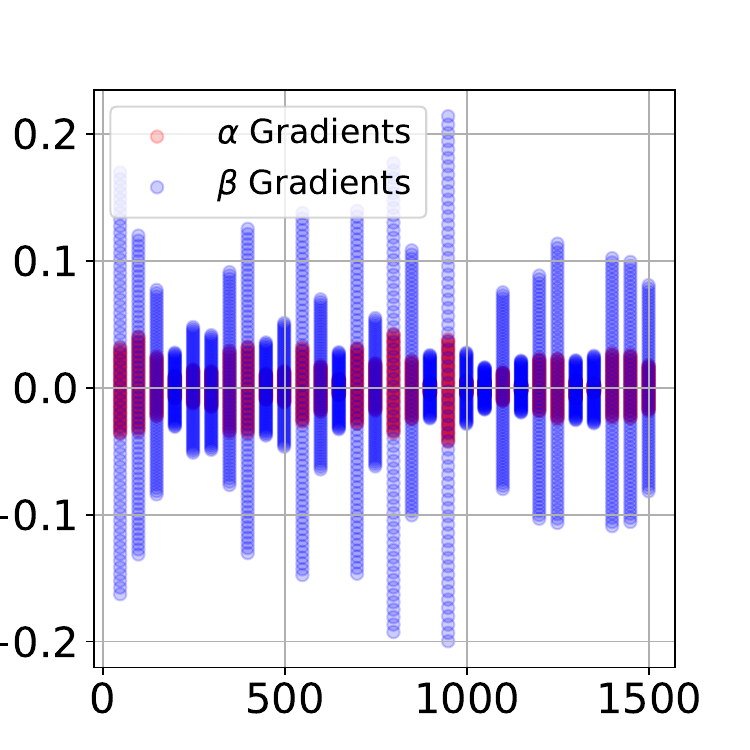}
		\caption{Distribution of gradient $\boldsymbol{\alpha}_t$ and $\boldsymbol{\beta}_t$.}
		\label{fig:wei:Grad:echo}
	\end{subfigure}
        \hspace{0.03mm}
	\begin{subfigure}{0.49\linewidth}
		\centering
		\includegraphics[width=1.0\linewidth]{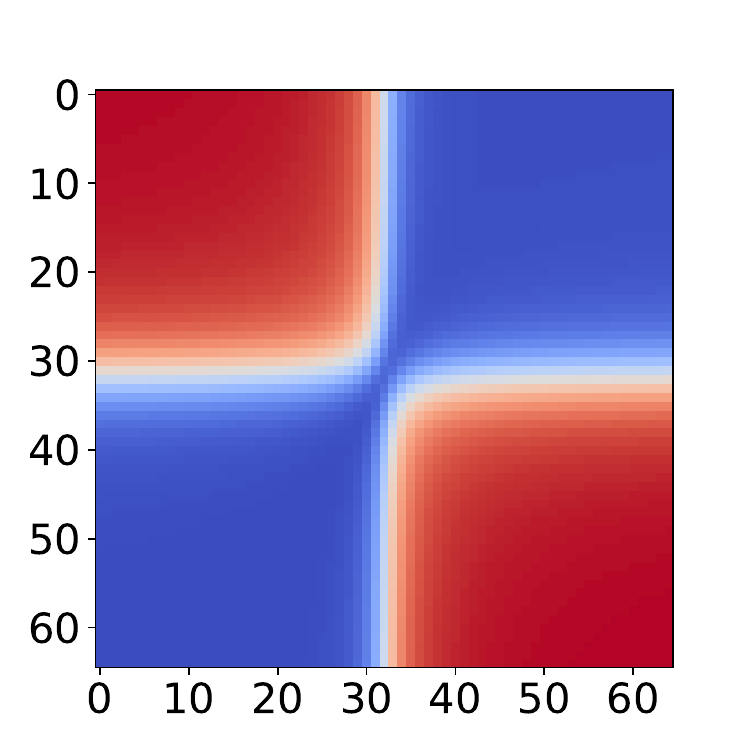}
		\caption{Pearson Correlation of gradient $\boldsymbol{\alpha}_t$ and $\boldsymbol{\beta}_t$. }
		\label{fig:wei:person:echo}
	\end{subfigure}
    \caption{Weights of parameters $\boldsymbol{\alpha}_t$ and $\boldsymbol{\beta}_t$ over training steps on EchoNet-Dynamic.}
    \label{fig:wei:echo}
\end{figure}

We also conducted experiments on the weights of parameters $\boldsymbol{\alpha}_t$ and $\boldsymbol{\beta}_t$ over training steps on the EchoNet-Dynamic dataset, as shown in \cref{fig:wei:echo}. 

These figures show how $\boldsymbol{\alpha}_t$ (controlling memory decay) and $\boldsymbol{\beta}_t$ (balancing old and new information) evolve over training. Their values fluctuate roughly between $-0.06$ and $+0.06$, and they often become more concentrated or exhibit recurring ``hot spots'' as training progresses. Such distributions suggest that the model continually adjusts these gating factors to align with changing task demands, eventually reaching relatively stable (or task-optimal) regions.

Although the gradients for these parameters frequently display positive or negative spikes, most gradient values remain within a stable range. This indicates that the network periodically makes significant adjustments to ``forget'' old information or to ``retain'' it when needed. Large gradient magnitudes can pose training challenges, highlighting the importance of proper learning rate settings, gradient clipping, or regularization.

Across different value ranges, the gradients of $\alpha$ and $\beta$ can correlate positively or negatively, forming visually distinct blocks or diagonal patterns. In certain ranges, an increase in $\alpha$ (faster memory decay) may coincide with an increase in $\beta$ (boosting new information), whereas in other ranges $\alpha$ and $\beta$ change in opposite directions. These patterns reflect how the network coordinates the interplay between forgetting past data and incorporating new inputs.

First, $\alpha$ and $\beta$ are learnable rather than fixed, undergoing notable shifts throughout training to accommodate evolving tasks. 
Second, their correlated gradients reveal how the model dynamically manages memory decay and integration to address varying environments or data patterns. 
Third, occasional extreme gradient values suggest training instability when balancing ``forget--retain'' operations, emphasizing the need for methods that mitigate abrupt parameter changes. 
If further experiments show that final values of $\alpha$ and $\beta$ strongly influence performance---for instance, by adopting larger $\alpha$ in scenarios requiring rapid forgetting---it would confirm that this learnable gating mechanism indeed supports the flexible discarding of outdated information.                
\end{document}